\documentclass[letterpaper, 10 pt, conference]{ieeeconf}  

\usepackage{xcolor}

\usepackage{graphicx} 
\usepackage{balance}
\usepackage{amsmath}
\usepackage{booktabs} 
\usepackage{array}   
\usepackage{multirow}
\usepackage{tabularx} 
\usepackage{ragged2e} 
\usepackage{siunitx}
\usepackage{amsfonts}
\usepackage{CJKutf8}
\usepackage{stfloats}
\usepackage{afterpage} 
\usepackage{float}

\IEEEoverridecommandlockouts                              

\title{\LARGE \bf
WaveComm: Lightweight Communication for Collaborative Perception via Wavelet Feature Distillation
}

\author{Erdemt Bao$^{1*\dagger}$, Jin Yang$^{2*\ddagger}$
\thanks{$^{1}$School of Mechanical Science and Engineering, Huazhong University of Science and Technology, Wuhan 430074, China.
        {\tt\small baoerdemt366@gmail.com}}%
\thanks{$^{2}$National Key Laboratory of Human-Machine Hybrid Augmented Intelligence, and Institute of Artificial Intelligence and Robotics, Xi'an Jiaotong University, Shaanxi 710049, China.
        {\tt\small jin.y.hust@gmail.com}}%
\thanks{*These authors contributed equally to this work.}
\thanks{$^{\dagger}$Corresponding author. $^\ddagger$Project Leader.}
\thanks{GitHub page: {\tt\small https://github.com/erdemtbao/WaveComm}}
}

\begin{document}

\maketitle
\thispagestyle{empty}
\pagestyle{empty}

\begin{abstract}

In multi-agent collaborative sensing systems, substantial communication overhead from information exchange significantly limits scalability and real-time performance, especially in bandwidth-constrained environments. This often results in degraded performance and reduced reliability. To address this challenge, we propose WaveComm, a wavelet-based communication framework that drastically reduces transmission loads while preserving sensing performance in low-bandwidth scenarios. The core innovation of WaveComm lies in decomposing feature maps using Discrete Wavelet Transform (DWT), transmitting only compact low-frequency components to minimize communication overhead. High-frequency details are omitted, and their effects are reconstructed at the receiver side using a lightweight generator. A Multi-Scale Distillation (MSD) Loss is employed to optimize the reconstruction quality across pixel, structural, semantic, and distributional levels. Experiments on the OPV2V and DAIR-V2X datasets for LiDAR-based and camera-based perception tasks demonstrate that WaveComm maintains state-of-the-art performance even when the communication volume is reduced to 86.3\% and 87.0\% of the original, respectively. Compared to existing approaches, WaveComm achieves competitive improvements in both communication efficiency and perception accuracy. Ablation studies further validate the effectiveness of its key components.
\end{abstract}

\section{INTRODUCTION}

Collaborative perception enables multiple agents to overcome inherent limitations of single viewpoints (e.g., occlusion, restricted sensing range) by exchanging complementary information through communication. This paradigm has been widely applied in connected and automated vehicles (CAVs), supported by Vehicle-to-Vehicle (V2V) and vehicle-to-infrastructure (V2I) technologies, significantly enhancing safety and efficiency in complex traffic environments \cite{arnold2020cooperative, wang2020v2vnet, liu2023towards}.

However, the benefits of collaboration come at the cost of substantial communication overhead. Raw sensor sharing or dense feature map exchange generates extremely high bandwidth demands, which are impractical in real-world deployments \cite{xu2023v2v4real, yu2022dair}. Even compression-oriented strategies (e.g., feature sparsification \cite{liu2020when2com}, learned transmission \cite{mao2024diffcp}) face trade-offs. They often lead to excessive bandwidth consumption under dense traffic, and transmit redundant or noisy information that does not contribute to downstream tasks, thereby reducing system efficiency.

\begin{figure}[t] 
  \centering
  \includegraphics[width=\columnwidth]{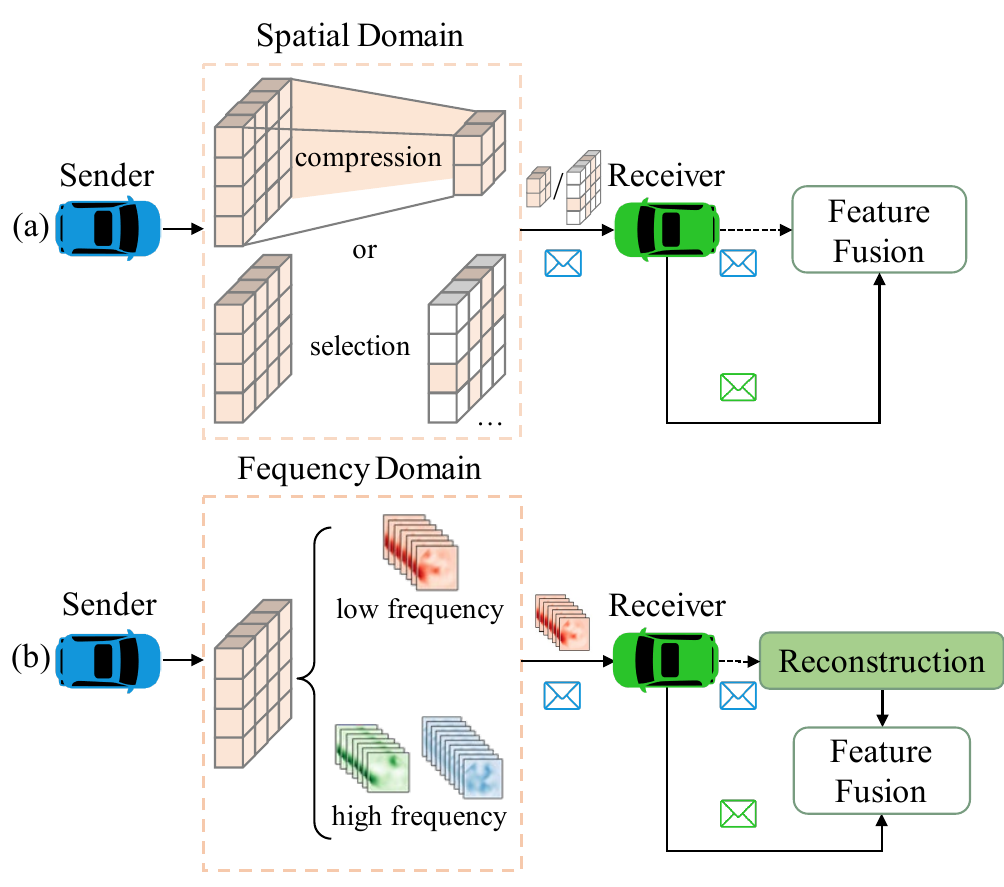} 

    \caption{Feature transmission methods in collaborative perception. (a) Spatial domain methods: The sender vehicle encodes data, applies compression or selection, and transmits it to the receiver vehicle, which performs feature fusion and decoding. (b) Frequency domain methods: The sender encodes data into low- and high-frequency components, primarily transmitting low-frequency data. The receiver reconstructs the original features from these components, followed by feature fusion and decoding.}
  \label{fig:introduction}
\end{figure}

A key limitation of existing approaches \cite{liu2020when2com,mao2024diffcp,xu2022v2x,zhang2024ermvp,jin2025bandwidth} is that they operate solely in the spatial domain, where redundancy is reduced by compressing or selecting feature maps, as shown in Figure~\ref{fig:introduction}(a).
While such strategies can reduce redundancy, they overlook the inherent frequency structure of perceptual signals. In the frequency domain, signals naturally decompose into complementary components: low-frequency bands that capture global semantics and structural contexts, and high-frequency bands that provide some details such as edges and textures. This inherent separation offers a principled way to design communication schemes that are both compact and information-preserving.

Based on this insight, we propose WaveComm, a wavelet-based collaborative perception framework that achieves communication efficiency by explicitly operating in the spatial frequency domain, similar in spirit to JPEG-style frequency-domain processing, as illustrated in Figure~\ref{fig:introduction}(b).
Specifically, WaveComm decomposes intermediate features using the Discrete Wavelet Transform (DWT) and transmits only the compact low-frequency components. Although high-frequency components encode edges, textures, and contour sharpness, their incremental contribution to collaborative perception is small relative to their communication cost. We therefore omit them from transmission and instead reconstruct their effects at the receiver side. At the receiver side, instead of applying the Inverse Discrete Wavelet Transform (IDWT) to recover the original features, we design a lightweight generator that reconstructs full feature maps directly from the transmitted low-frequency components. The generator is optimized using a combination of reconstruction loss, perceptual loss, structural similarity loss, and adversarial loss, ensuring that the reconstructed features remain both semantically consistent and task-aligned.

Through the above design, WaveComm realizes a principled frequency-aware communication scheme that reduces bandwidth while preserving task-critical semantics. Moreover, the decomposition into low-frequency components is compatible with existing compression and feature selection strategies, making it easy to integrate WaveComm with prior frameworks for further improvements. Furthermore, we extend our design with multi-level wavelet decomposition, which enables progressive transmission and reconstruction at different resolutions, further enhancing flexibility under varying bandwidth constraints.

We summarize the contributions of this work as follows:
\begin{itemize}
\item We introduce a wavelet-based framework that reduces communication cost by selectively transmitting low-frequency components while retaining sufficient information for downstream tasks.

\item We design a Wavelet Feature Distillation Module to reconstruct full features from transmitted low-frequency components and optimize it with a hybrid objective.

\item Extensive experiments show that WaveComm achieves a superior efficiency–accuracy trade-off. Ablation studies further confirm the effectiveness of its modules.
\end{itemize}

\section{RELATED WORK}

\subsection{Communication-Efficient Cooperative Perception}
To reduce communication costs, a variety of communication-efficient methods have recently been proposed.
Methods like Where2comm \cite{hu2022where2comm} and CodeFilling \cite{hu2024communication} selectively transmit critical features using spatial confidence or codebook representations, balancing efficiency and performance, though requiring prior single-agent perception that increases latency. CMiMC \cite{su2024makes} maximizes mutual information to preserve discriminative data. How2Comm \cite{yang2023how2comm}, ERMVP \cite{zhang2024ermvp}, and FFNet \cite{yu2023flow} exploit spatial sparsity of foreground objects or temporal correlations for downsampling and flow-based transmission. Fast2comm \cite{zhang2025fast2comm} uses bounding box priors to minimize noise and adapt to localization errors.
DiscoNet \cite{li2021learning} applies matrix-valued weights and teacher-student frameworks for interactions. Transformer methods include V2X-ViT \cite{xu2022v2x} for V2X cooperation, CoBEVT \cite{xu2022cobevt} for BEV segmentation, and HM-ViT \cite{xiang2023hm} with sparse heterogeneous attentions for multi-agent multi-camera 3D detection and hetero-modal perception. Recent approaches like CoCMT \cite{wang2025cocmt} and CoopDETR \cite{wang2025coopdetr} leverage object queries for efficient transmission, while Which2comm \cite{yu2025which2comm} integrates semantic detection boxes for sparse, object-level sharing.

Current research on communication-efficient cooperative perception focuses heavily on spatial domain techniques, often overlooking the frequency domain's potential. Frequency-based methods allow precise control over data transmission, prioritizing key components to reduce redundancy while improving computational efficiency.

\subsection{Wavelet Transforms in Feature Processing}
Wavelet transforms are widely applied in image processing to enhance visual quality, computational efficiency, and feature representation. In generative models, SWAGAN \cite{gal2021swagan} integrates wavelets into GANs to improve image quality and performance, while Wavelet-srnet \cite{huang2017wavelet} predicts wavelet coefficients for super-resolution image reconstruction. Wavelet-based methods like CWNN-MRF \cite{duan2017sar} and Wavelet-Pooling \cite{williams2018wavelet} replace pooling operators in CNNs, using dual-tree complex wavelet transforms or learned wavelet bases to enhance performance without compression.
For compression, wavelet-based approaches optimize neural network efficiency. Efficient Wavelet-Based Linear Layers \cite{wolter2020neural} learn wavelet basis functions to compress linear layer weights, unlike activation compression. WCC \cite{finder2022wavelet} uses Haar-wavelet transforms to compress feature maps, integrating with point-wise convolutions to reduce costs in image-to-image tasks. Masked Wavelet NeRF \cite{rho2023masked} applies wavelets to grid-based neural fields for efficient parameter compression, maintaining data structure benefits. HL-RSCompNet \cite{xiang2024remote} employs Discrete Wavelet Transform (DWT) to split features into high- and low-frequency components, enhancing compression via frequency domain encoding-decoding. Similarly, UGDiff \cite{song2024high} uses wavelets for high-frequency compression in diffusion models, predicting high frequencies and compressing residuals to improve fidelity.

Despite these advances, the use of frequency domain wavelet transforms in multi-agent collaborative perception remains limited. To address this gap, this work explores feature processing in the frequency domain and proposes a novel decomposition and reconstruction mechanism.

\section{PROBLEM FORMULATION}

We investigate a cooperative perception framework comprising $N$ agents, each undertaking distinct detection tasks. In this setup, each agent simultaneously serves as a contributor, sharing perceptual data with others, and as a recipient, leveraging data received from peers. Let $\mathcal{X}_i$ denote the sensory input (e.g., from LiDAR or cameras) collected by the $i$-th agent, and let $\mathcal{G}^0_i$ represent the corresponding ground-truth annotations for 3D object detection. The goal is to optimize the detection model parameters to maximize the aggregate detection performance, subject to a total communication budget $C$. Specifically, the optimization problem is formulated as:

\begin{equation}
\begin{aligned}
\underset{\theta,\mathcal{P}}{\text{argmax}} \sum_{i=1}^N h \left( \Phi_{\theta} \left( \mathcal{X}_i, \{ \mathcal{P}_{j \to i} \}_{j=1}^N \right), \mathcal{G}^0_i \right),  \\
\text{s.t.} \quad \sum_{\substack{i,j=1 \\ j \neq i}}^N c(\mathcal{P}_{j \to i}) \leq C \label{eq:setup}
\end{aligned}
\end{equation}
where $\Phi_{\theta}$ is the 3D object detection model parameterized by $\theta$, $\mathcal{P}_{j \to i}$ represents the message transmitted from agent $j$ to agent $i$, $h(\cdot, \cdot)$ is the evaluation metric for detection performance, and $c(\cdot)$ quantifies the communication cost of the messages. The primary challenge lies in designing the messages $\mathcal{P}_{j \to i}$ to be both informative for enhancing detection and compact to satisfy the communication constraint.

\textbf{Communication volume.} We follow the communication volume definition from Where2comm \cite{hu2022where2comm}, but use float32 or float16 instead of a fixed 32-bit representation. For the message sent from the $i$-th to the $j$-th agent, the binary selection matrix $\mathbf{M}_{i \to j} \in \mathbb{R}^{H \times W}$ represents the spatial grid, with $H$ and $W$ as height and width. The communication volume is:

\begin{equation}
\log_2 \left( |\mathbf{M}_{i \to j}| \times D \times n_L/8 \right)
\label{eq:communication_volume}
\end{equation}
where $\left| \cdot \right|$ denotes the L0 norm, indicating the count of non-zero entries in the binary selection matrix (i.e., the total spatial grids transmitted), $D$ represents the channel dimension, $n_L$ corresponds to the float32 or float16 data type, and dividing by 8 converts the result to bytes.

\section{METHODOLOGY}
\subsection{Overview of WaveComm Architecture}

\begin{figure*}[htbp]
  \centering
  \includegraphics[width=\textwidth]{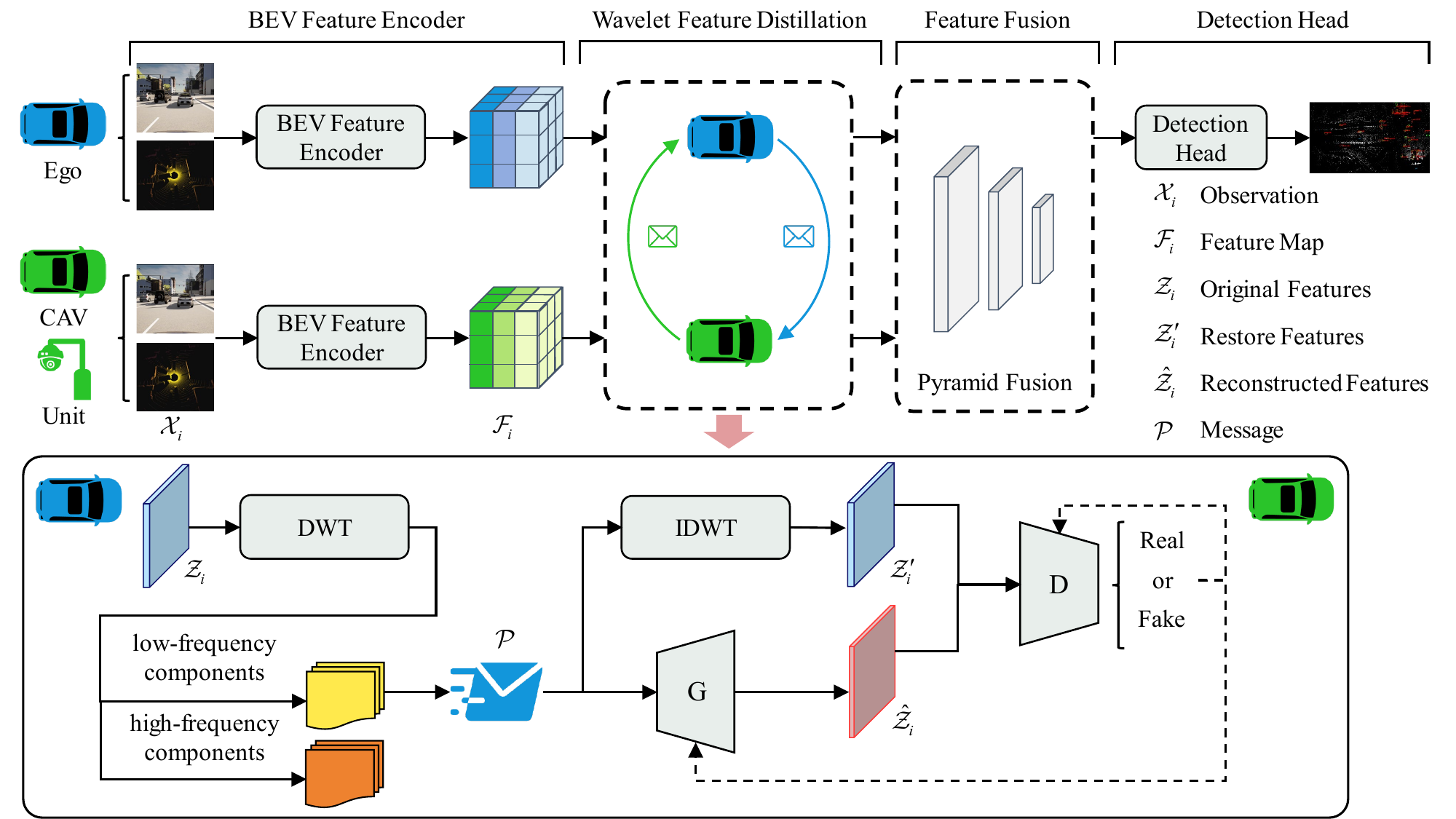}   
  \caption{Overview of WaveComm. WaveComm enables efficient information exchange among intelligent agents to support collaborative autonomous driving. (a) BEV Feature Encoder, which converts agent observations into BEV feature maps.
(b) Wavelet Feature Distillation, which employs DWT to decompose features into low- and high-frequency components, followed by a Wavelet Generator and Wavelet Discriminator for efficient feature reconstruction using IDWT.
(c) Feature Fusion, which integrates features from multiple agents to enhance the overall feature representation effectively.
(d) Detection Head, which produces final detection outputs based on the fused features.}
  \label{fig:framework}
\end{figure*}

The architecture of WaveComm is shown in Figure~\ref{fig:framework}. Blue represents the ego vehicle, green denotes collaborative Connected Autonomous Vehicles (CAVs) or Road-Side Units (RSUs). Multiple agents uniformly process their individual observations, denoted as $\mathcal{X}_i$, through a Bird's-Eye-View (BEV) feature encoder to extract corresponding feature maps $\mathcal{F}_i$. These feature maps $\mathcal{F}_i$ are further processed to generate the target features $\mathcal{Z}_i$ for transmission. The features $\mathcal{Z}_i$ are decomposed using DWT into low-frequency and high-frequency components. To enhance communication efficiency in collaborative perception, only the low-frequency components are transmitted across the collaboration link. 

At the receiving end, the Wavelet Feature Distillation module reconstructs the original $\mathcal{Z}_i$ by recovering the missing high-frequency details. This process begins with the compressed low-frequency component processed through IDWT to yield restored features $\mathcal{Z}'_i$, which serve as a supervisory signal. Next, a Generator module, using data priors and adversarial learning with a Discriminator, produces reconstructed features $\hat{\mathcal{Z}}_i$. The Discriminator, combined with a Multi-Scale Distillation (MSD) Loss, evaluates the authenticity of the reconstructed features, ensuring high-fidelity reconstruction and close resemblance to the original features.
All reconstructed $\hat{\mathcal{Z}}_i$ from participating agents are then fused by the Pyramid Fusion network \cite{lu2024extensible}. During feature fusion, each agent’s reconstructed BEV feature $\hat{\mathcal{Z}}_i$ is warped into the ego coordinate frame using pairwise affine transformations, and a per-cell softmax-weighted sum across agents is computed to obtain the fused BEV feature, which is finally fed into the detection head.

\subsection{BEV Feature Encoder}

For each agent, the input data $\mathcal{X}_i$, which may consist of RGB images or 3D point clouds, is converted into a BEV feature representation. This method allows all agents to project their individual sensory inputs into a unified global coordinate system, streamlining collaboration by removing the need for intricate coordinate transformations. The BEV encoder processes the input to produce a feature map $\mathcal{F}_i \in \mathbb{R}^{H \times W \times C}$, where $H$, $W$, and $C$ denote the height, width, and number of channels, respectively. By sharing a common BEV coordinate system, agents can efficiently integrate and process data from either RGB images or point clouds, fostering seamless multi-agent cooperation. We use the term “feature” to refer to these learned CNN feature maps rather than raw occupancy counts, and in this work the BEV features are computed per frame without temporal aggregation.

\subsection{Wavelet Feature Distillation Module}

The Wavelet Feature Distillation Module consists of two parts: feature decomposition and feature reconstruction.
We use DWT for feature decomposition, breaking down BEV features into low-frequency and high-frequency components. Low-frequency components retain most semantic information and global structure, and are the core basis for perception tasks. High-frequency components, on the other hand, primarily contain local details such as edges, textures, and contours. While they offer limited gains in perception accuracy, they incur additional communication overhead. Therefore, we only transmit low-frequency components during inter-vehicle communication to reduce bandwidth and computing resource consumption. Distinct from naive spatial downsampling which uniformly discards information, DWT preserves the global spatial structure while removing high-frequency details.

\begin{figure}[htbp]  
  \centering
  \includegraphics[width=\columnwidth]{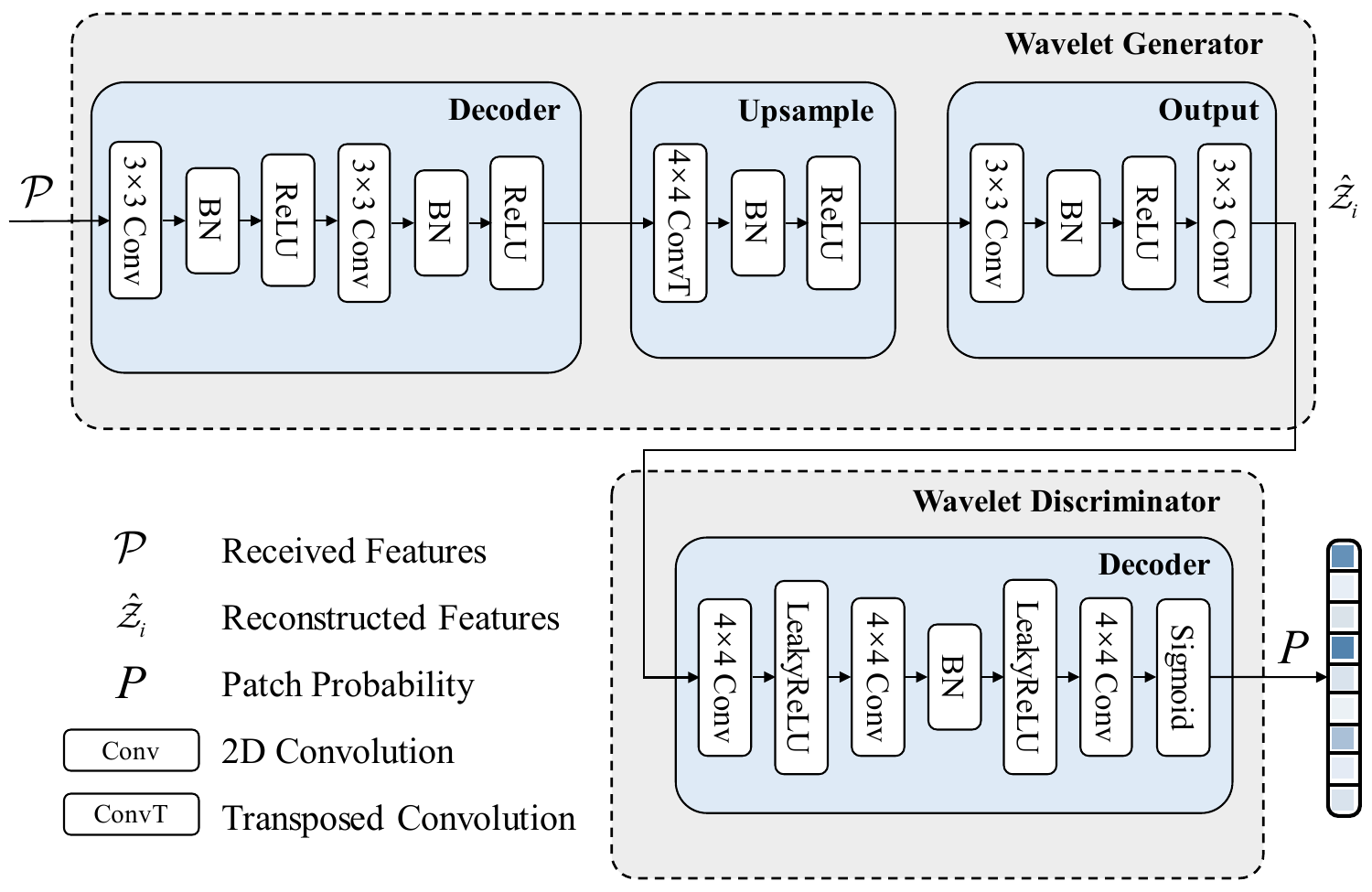} 
  \caption{Architecture of Wavelet Generator and Wavelet Discriminator. The Wavelet Generator uses Decoder, Upsample, and Output modules to reconstructed features $\hat{\mathcal{Z}}_i$ from the transmitted low-frequency component. The Wavelet Discriminator employs Sigmoid to generate a probability map $P$ like PatchGAN \cite{isola2017image}.}
  \label{fig:G_D}
\end{figure}

For feature reconstruction, the most direct approach is to use IDWT. However, since missing high-frequency components must be zeroed out, the resulting features are often overly smooth and fail to meet the discriminative feature requirements of downstream detection networks. To address this, we propose a Wavelet Generator that directly generates complete features from low-frequency components. Using data priors to infer missing high-frequency information, the resulting features are optimized through a series of loss functions. This ensures that the generated features maintain global consistency while recovering richer details and semantics, effectively improving downstream perception performance while ensuring communication efficiency. 
The Wavelet Generator is divided into three main modules: the Decoder, Upsample, and Output stages, as illustrated in Figure~\ref{fig:G_D}. The Decoder consists of two $3\times3$ convolutional layers with batch normalization (BN) and ReLU activation, processing the input low-frequency component into intermediate features. The Upsample module employs a $4\times4$ transposed convolutional layer (ConvTranspose) with stride 2, BN, and ReLU to upscale the features, doubling the spatial dimensions. The Output stage includes two additional $3\times3$ convolutional layers with BN and ReLU, producing the reconstructed features $\hat{\mathcal{Z}}_i$.

In addition, we further introduce the Wavelet Discriminator, which constrains the generator output through adversarial learning, making it not only globally consistent with the low-frequency input but also closer to the true features in terms of distribution. Specifically, the discriminator is trained to distinguish features recovered by the generator from true features, while the generator tries to fool the discriminator by minimizing the adversarial loss, thereby learning a sharper and more semantically consistent representation. The Wavelet Discriminator comprises a single Decoder module with three $4\times4$ convolutional layers, each with stride 2, interspersed with LeakyReLU activation and BN, as illustrated in Figure~\ref{fig:G_D}, culminating in a Sigmoid activation to output a probability map $P$ for authenticity assessment.

\subsection{Multi-Scale Distillation Loss}

During training, we use features derived from low-frequency components via an IDWT as the generator's supervisory signal. While the IDWT reconstruction results appear too smooth due to the lack of high-frequency components and are therefore unsuitable as the final output for direct perception tasks, they are physically identical to the transmitted low-frequency components, providing a stable and reasonable learning target for the generator. Directly using the original features as supervision would force the model to recover high-frequency information that was never transmitted during communication, making the training objective unattainable. In contrast, the IDWT results provide the generator with a baseline that complies with frequency domain constraints. While maintaining low-frequency consistency, the generator further recovers detailed and discriminative features through Multi-
Scale Distillation (MSD) Loss combined with the following four levels of loss, thereby reconstructing features that are both physically reasonable and useful for downstream tasks.

\begin{itemize}

\item Pixel-Level, Reconstruction Loss: Ensures element-by-element numerical similarity through L1 loss.
\item Structural-Level, SSIM Loss: Focuses on structural, brightness, and contrast similarity of feature maps through SSIM loss.
\item Semantic-Level, Perceptual Loss: Emphasizes high-level semantic similarity through perceptual loss.
\item Distributional-Level, Adversarial Loss: Ensures that the distribution of the generated features is close to the real features through adversarial loss.

\end{itemize}

The “multi-scale” nature of the MSD loss is evident in its hierarchical constraints across low- to high-level features, potentially supporting applications such as feature compression and cooperative perception.

For a given agent, we denote its IDWT-restored features $\mathcal{Z}'$ 
as $f$ and its generator-reconstructed features $\hat{\mathcal{Z}}$ 
as $\hat{f}$ for brevity. The subscript $n$ indexes individual spatial elements after flattening these feature tensors, with $N$ denoting the total number of elements.

\textbf{Reconstruction Loss.}
The low-frequency component of the wavelet decomposition is reconstructed into the original information to recover one of its original representations. The element-by-element difference between the reconstructed features and the restored features is measured using the L1 loss, which encourages accurate reconstruction at the pixel level, with the reconstructed features being as close as possible to the restored features in terms of value.

\begin{equation}
\mathcal{L}_{\text{Recon}} = \frac{1}{N} \sum_{n=1}^N \left| \hat{f}_n - f_n \right|
\end{equation}

\textbf{Perceptual Loss.}
During the reconstruction process, the goal is to ensure that the reconstructed features closely resembles the restored features in terms of high-level semantic content. The perceptual loss measures the feature similarity by computing the mean square error (MSE) between the normalized reconstructed features and the normalized restored features. This encourages semantic consistency in the high-dimensional feature space, beyond mere pixel-level accuracy. The perceptual loss function is defined as:

\begin{equation}
\mathcal{L}_{\text{Percep}} = \frac{1}{N} \sum_{n=1}^N \left( \frac{\hat{f}_n}{\|\hat{f}\|_2} - \frac{f_n}{\|f\|_2} \right)^2
\end{equation}
where:
\begin{itemize}
    \item $\|\hat{f}\|_2$, $\|f\|_2$: The L2 norm of the reconstructed and restored feature vectors, respectively, used for normalization.
    \item $\frac{\hat{f}_n}{\|\hat{f}\|_2}$, $\frac{f_n}{\|f\|_2}$: The normalized $n$-th elements of the reconstructed and restored features, respectively.
\end{itemize}

\textbf{Structural Similarity Loss.}
This loss calculates the structural similarity index (SSIM), which is used to measure the similarity between two feature maps in terms of brightness, contrast, and structure.
The structural similarity of feature maps is emphasized to make up for the insensitivity of L1 loss to structural information.

\begin{equation}
\mathcal{L}_{\text{SSIM}} = 1 - \text{SSIM}(\hat{f}, f) 
\end{equation}

\textbf{Adversarial Loss.}
The generator creates reconstructed features to mimic restored features, aiming to fool the discriminator. The discriminator distinguishes restored features from reconstructed ones, maximizing classification accuracy. The discriminator loss enhances this accuracy, while the generator loss optimizes the generator to produce reconstructed features indistinguishable from restored features.

\begin{equation}
\begin{aligned}
\mathcal{L}_{\text{D}} &= \text{BCE}(D(f), 1) + \text{BCE}(D(\hat{f}), 0) \\
&= -\frac{1}{N} \sum_{n=1}^N \left[ \log(D(f_n)) + \log(1 - D(\hat{f}_n)) \right]
\end{aligned}
\end{equation}
\begin{equation}
\mathcal{L}_{\text{G}} = \text{BCE}(D(\hat{f}), 1) = -\frac{1}{N} \sum_{n=1}^N \log(D(\hat{f}_n))
\end{equation}
where:
\begin{itemize}
    \item $D(\cdot)$: The discriminator function, outputting a probability that the input feature is real.
    \item $\text{BCE}$: The binary cross-entropy loss function.
\end{itemize}

\textbf{Total Loss.}
The total loss is obtained by taking the weighted sum of the aforementioned losses.

\begin{equation}
\begin{aligned}
\mathcal{L}_{\text{MSD}} &= 
\begin{cases}
\mathcal{L}_{\text{ReconTotal}} &= \lambda_{\text{recon}} \cdot \big( \alpha \cdot \mathcal{L}_{\text{Recon}} + \beta \cdot \mathcal{L}_{\text{SSIM}} \\
&\quad + \gamma \cdot \mathcal{L}_{\text{Percep}} \big) \\
\mathcal{L}_{\text{Adv}} &= \lambda_{\text{adv}} \cdot \mathcal{L}_{\text{G}} \\
\mathcal{L}_{\text{D}} &= \text{BCE}(D(f), 1) + \text{BCE}(D(\hat{f}), 0)
\end{cases}
\end{aligned}
\end{equation}
where:
\begin{itemize}
    \item $\mathcal{L}_{\text{Recon}}$, $\mathcal{L}_{\text{SSIM}}$, $\mathcal{L}_{\text{Percep}}$: The reconstruction, structural similarity, and perceptual loss components, respectively.
    \item $\mathcal{L}_{\text{G}}$, $\mathcal{L}_{\text{D}}$: The generator and discriminator losses in the adversarial framework, respectively.
    \item $\lambda_{\text{recon}}$, $\lambda_{\text{adv}}$, $\alpha$, $\beta$, $\gamma$: Weighting coefficients to balance the contributions of each loss term.
\end{itemize}

\section{EXPERIMENTAL RESULTS}
\subsection{Dataset and Evaluation Metrics}

\textbf{Datasets.}
OPV2V \cite{xu2022opv2v} is a simulated dataset designed for V2V collaborative perception. It is constructed using the CARLA simulator \cite{dosovitskiy2017carla} and the OpenCDA framework \cite{xu2021opencda}, generating diverse driving scenarios with a particular emphasis on V2V communication for perception tasks. The dataset comprises 12,000 frames spanning eight different towns in CARLA and a digital replica of Culver City, Los Angeles, containing over 232,913 3D vehicle bounding boxes in total. It is designed to replicate complex real-world driving situations, including varying traffic densities and dynamic driving behaviors.
In contrast, DAIR-V2X \cite{yu2022dair} is a real-world collaborative perception dataset consisting of 9,000 frames. Each frame includes synchronized data from one vehicle and one RSU, both equipped with a LiDAR sensor and a 1920×1080 camera. The LiDAR on the RSU has 300 channels, while the vehicle-mounted LiDAR is 40-channel.

\textbf{Evaluation Metrics.}
We use Average Precision (AP) at Intersection-over-Union (IoU) thresholds of 0.3, 0.5, and 0.7 to evaluate 3D object detection performance. To evaluate the transmission cost, we follow the definition of communication volume in Eq.~\eqref{eq:communication_volume}, which calculates the communication volume by measuring the message size in bytes, expressed in a logarithmic scale with base 2. In other methods, $n_L$ is 32, while in our method, $n_L$ is 16.

\subsection{Experimental Setup}

\begin{table*}[htbp]
\centering
\caption{Performance Comparison on OPV2V and DAIR-V2X. Comm
 denotes the communication volume calculated with Eq.~\eqref{eq:communication_volume}.}
\label{tab:performance}
\small 
\resizebox{\textwidth}{!}{%
\begin{tabular}{l|cccccc|cccccc}
\toprule
\multicolumn{1}{c|}{Dataset} & \multicolumn{6}{c|}{OPV2V} & \multicolumn{6}{c}{DAIR-V2X} \\ \midrule
\multicolumn{1}{c|}{\multirow{2}{*}{Method}} & \multicolumn{3}{c|}{Camera-based} & \multicolumn{3}{c|}{LiDAR-based} & \multicolumn{3}{c|}{Camera-based} & \multicolumn{3}{c}{LiDAR-based} \\
\multicolumn{1}{c|}{} & AP50 $\uparrow$ & AP70 $\uparrow$ &\multicolumn{1}{c|}{Comm} & AP50 $\uparrow$ & AP70 $\uparrow$ & \multicolumn{1}{c|}{Comm} & AP30 $\uparrow$ & AP50 $\uparrow$ & \multicolumn{1}{c|}{Comm} & AP30 $\uparrow$ & \multicolumn{1}{c}{AP50 $\uparrow$} & Comm\\ \midrule
No Collaboration & 0.405 & 0.216 & \multicolumn{1}{c|}{0.0} & 0.782 & 0.634 & \multicolumn{1}{c|}{0.0} & 0.014 & 0.004 & \multicolumn{1}{c|}{0.0} & 0.421 & \multicolumn{1}{c}{0.405} & 0.0\\ \midrule
F-Cooper (2019) & 0.469 & 0.219 & \multicolumn{1}{c|}{22.0} & 0.763 & 0.481 & \multicolumn{1}{c|}{24.0} & 0.115 & 0.026 & \multicolumn{1}{c|}{23.0} & 0.723 & \multicolumn{1}{c}{0.620} & 23.0\\
DiscoNet (2021) & 0.517 & 0.234 & \multicolumn{1}{c|}{22.0} & 0.882 & 0.737 & \multicolumn{1}{c|}{24.0} & 0.083 & 0.017 & \multicolumn{1}{c|}{23.0} & 0.746 & \multicolumn{1}{c}{0.685} & 23.0\\
AttFusion (2022) & 0.529 & 0.252 & \multicolumn{1}{c|}{22.0} & 0.878 & 0.751 & \multicolumn{1}{c|}{24.0} & 0.094 & 0.021 & \multicolumn{1}{c|}{23.0} & 0.738 & \multicolumn{1}{c}{0.673} & 23.0\\
V2X-ViT (2022) & 0.603 & 0.289 & \multicolumn{1}{c|}{22.0} & 0.917 & 0.790 & \multicolumn{1}{c|}{24.0} & 0.198 & 0.057 & \multicolumn{1}{c|}{23.0} & 0.785 & \multicolumn{1}{c}{0.521} & 23.0\\
CoBEVT (2022) & 0.571 & 0.261 & \multicolumn{1}{c|}{22.0} & 0.935 & 0.821 & \multicolumn{1}{c|}{24.0} & 0.182 & 0.042 & \multicolumn{1}{c|}{23.0} & 0.787 & \multicolumn{1}{c}{0.692} & 23.0\\
HM-ViT (2023) & 0.643 & 0.370 & \multicolumn{1}{c|}{22.0} & 0.950 & 0.873 & \multicolumn{1}{c|}{24.0} & 0.163 & 0.044 & \multicolumn{1}{c|}{23.0} & 0.818 & \multicolumn{1}{c}{0.761} & 23.0\\ 
WaveComm (ours) & \textbf{0.681} & \textbf{0.451} & \multicolumn{1}{c|}{\textbf{19.0}} & \textbf{0.965} & \textbf{0.926} & \multicolumn{1}{c|}{\textbf{21.0}} & \textbf{0.274} & \textbf{0.123} & \multicolumn{1}{c|}{\textbf{20.0}} & \textbf{0.831} & \multicolumn{1}{c}{\textbf{0.790}} & \textbf{20.0}\\ \bottomrule
\end{tabular}%
}
\end{table*}

The experiments were conducted using the OpenCOOD framework on the OPV2V and DAIR-V2X datasets. 

For the LiDAR-based setup, we employ PointPillars \cite{lang2019pointpillars} as the feature encoder, which takes 64-channel LiDAR data as input. The detection range is set to $x, y \in [-102.4\text{m}, +102.4\text{m}]$. The feature map is downsampled by a factor of 2 and reduced to 64 dimensions to enable efficient communication. The multi-scale feature dimensions used in Pyramid Fusion are [64, 128, 256]. For the camera-based setup, the detection range is defined as $x, y \in [-51.2\text{m}, 51.2\text{m}]$. The BEV feature map is subsequently downsampled by a factor of 2 and compressed to 64 dimensions to facilitate message transmission. The parameters of MSD Loss are $\alpha=1.0,\ \beta=1.0,\ \gamma=0.1$.

For OPV2V, we trained end-to-end for 30 epochs using the Adam optimizer (learning rate 0.002) with multistep learning rate scheduling, a batch size of 1, on 8 NVIDIA GeForce RTX 4090 GPUs. For DAIR-V2X, we trained for 40 epochs with the same optimizer and scheduler, using a batch size of 2 on 8 RTX 4090 GPUs.

\subsection{Quantitative Results}

\begin{figure*}[htbp]
    \centering
    \begin{minipage}[b]{0.24\textwidth}
        \centering
        \includegraphics[width=\textwidth]{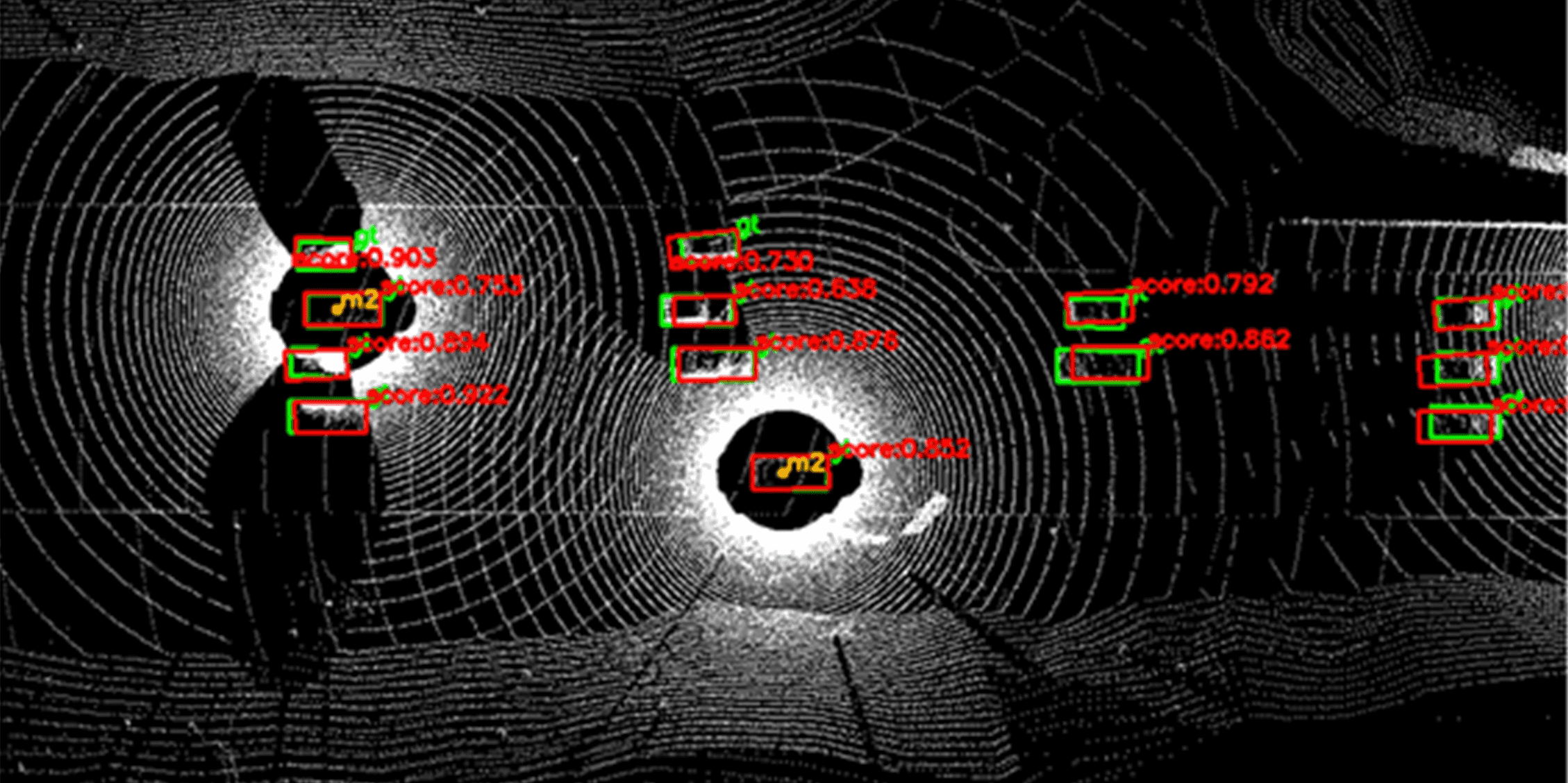}
    \end{minipage}%
    \hspace{2mm}%
    \begin{minipage}[b]{0.24\textwidth}
        \centering
        \includegraphics[width=\textwidth]{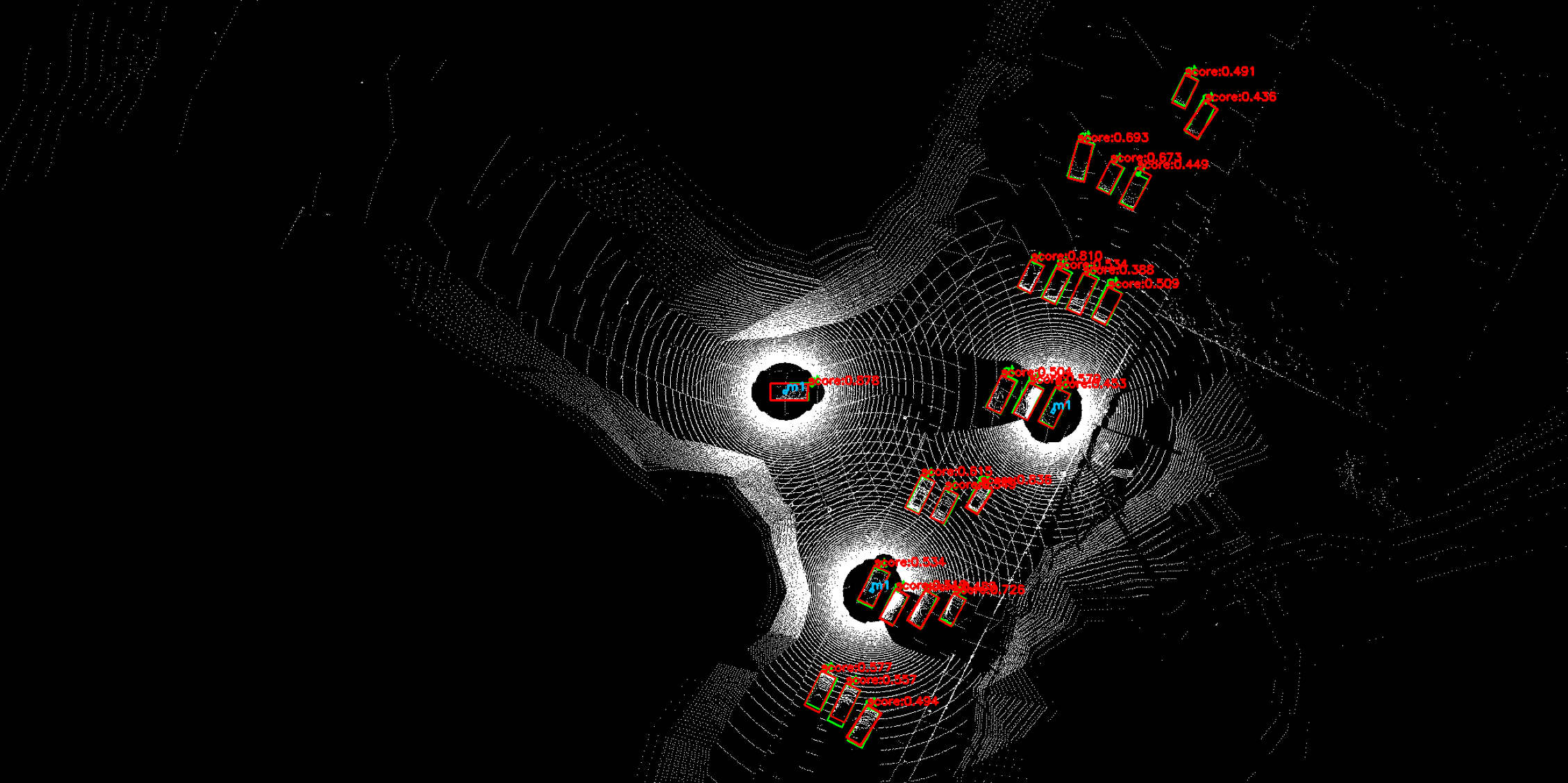}
    \end{minipage}%
    \hspace{2mm}%
    \begin{minipage}[b]{0.24\textwidth}
        \centering
        \includegraphics[width=\textwidth]{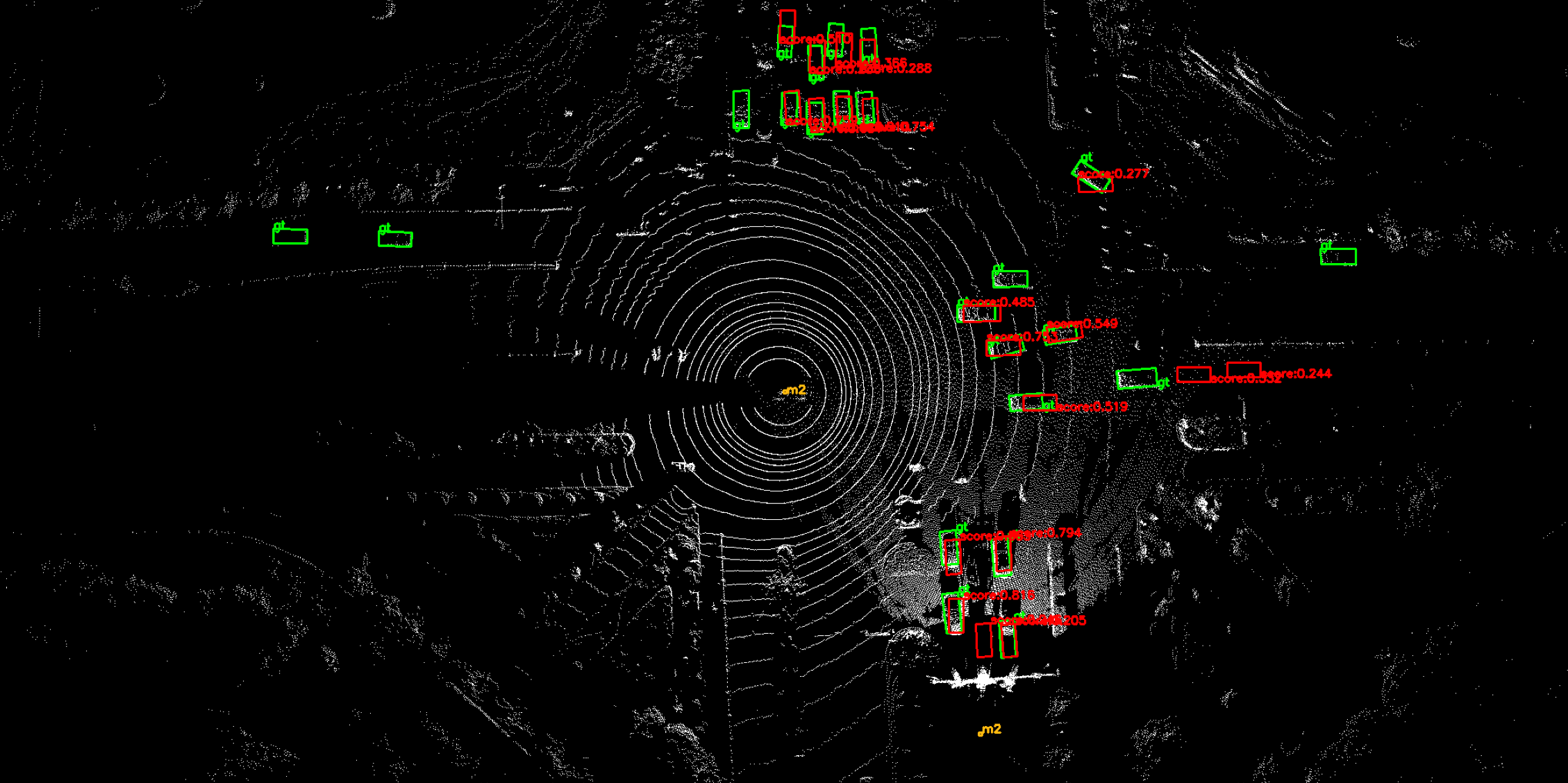}
    \end{minipage}%
    \hspace{2mm}%
    \begin{minipage}[b]{0.24\textwidth}
        \centering
        \includegraphics[width=\textwidth]{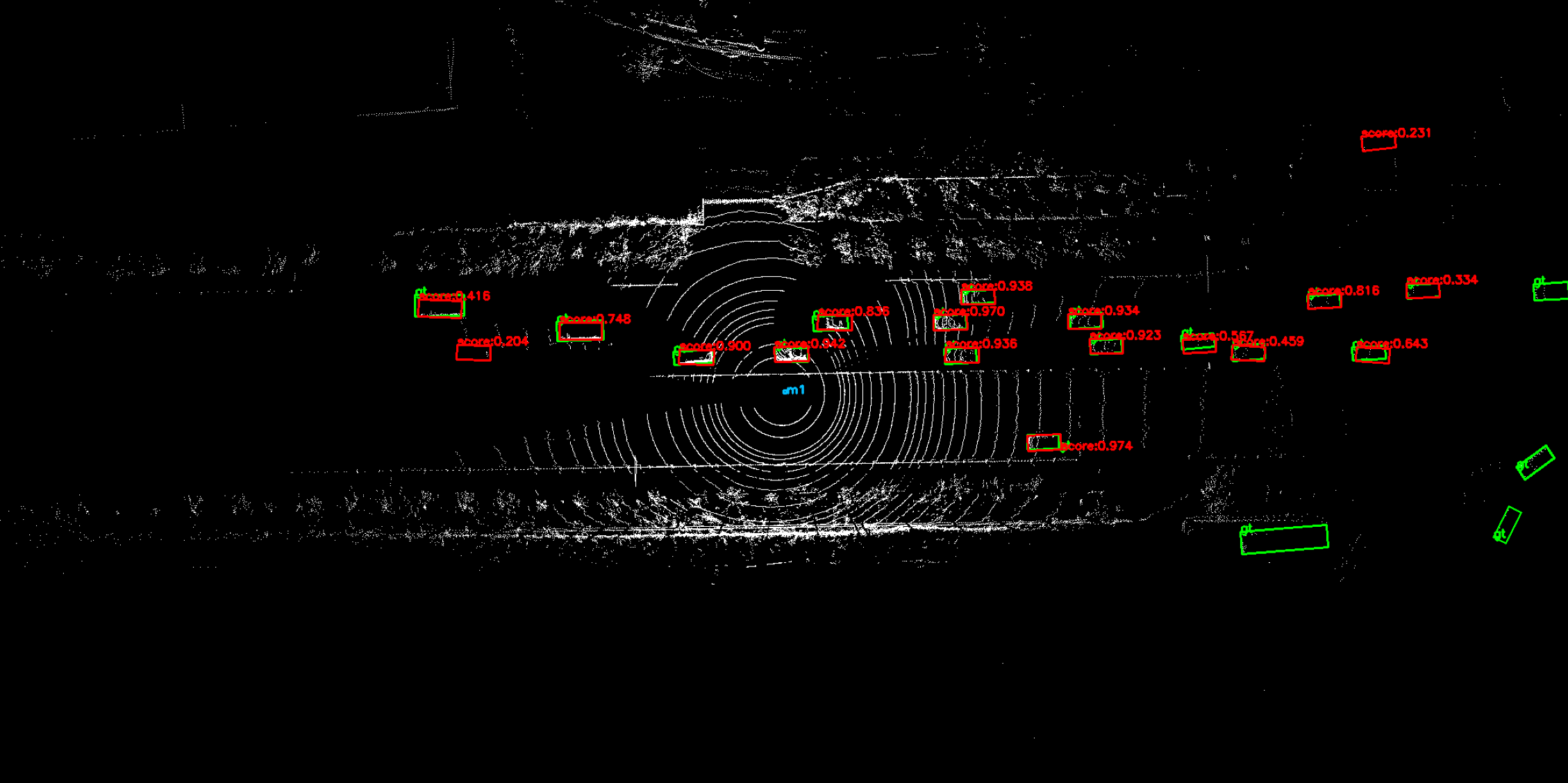}
    \end{minipage}%

    \vspace{0.4em} 

    \begin{minipage}[b]{0.24\textwidth}
        \centering
        \includegraphics[width=\textwidth]{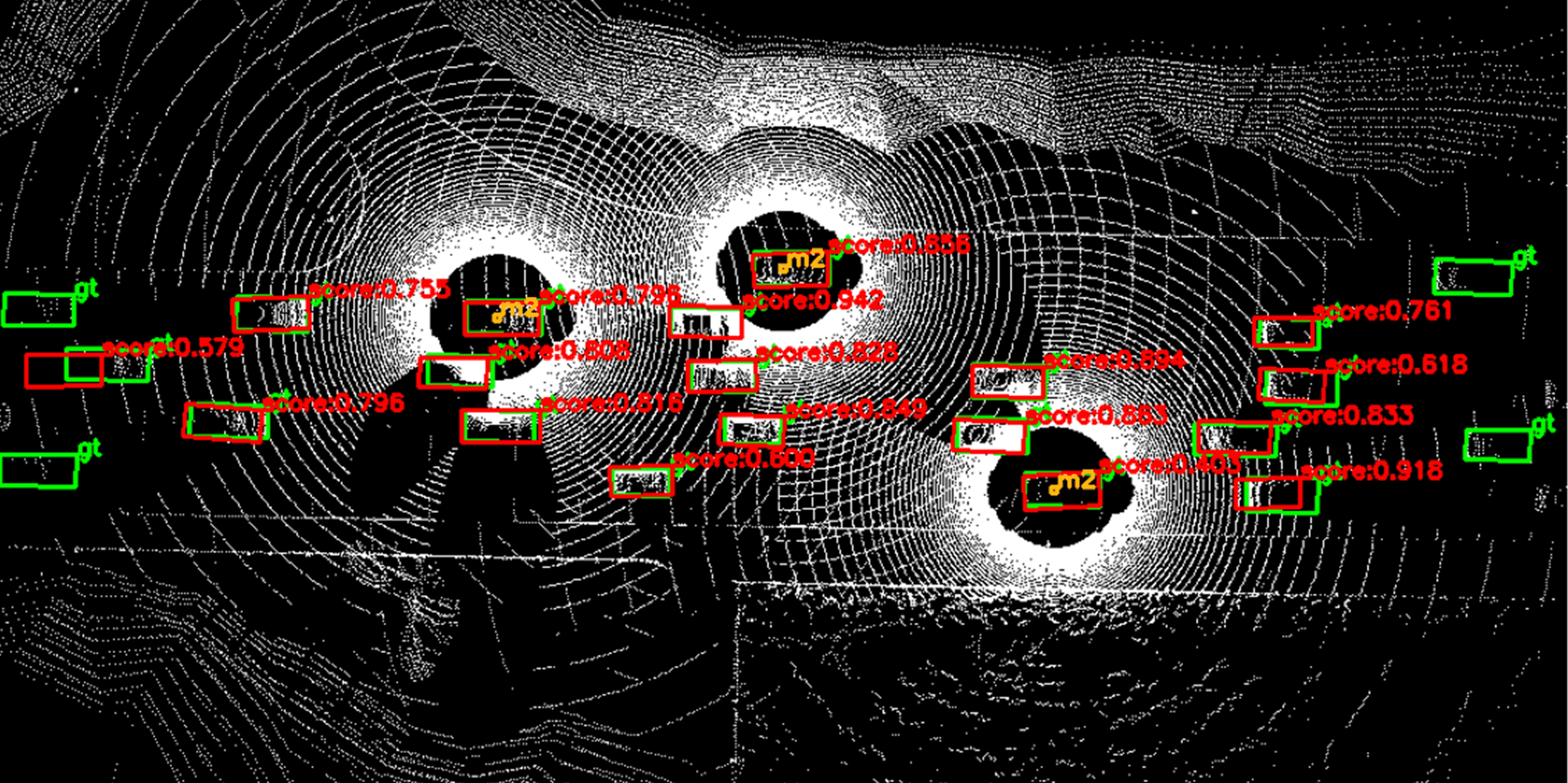}
        \smallskip
        \small OPV2V-Camera
    \end{minipage}%
    \hspace{2mm}%
    \begin{minipage}[b]{0.24\textwidth}
        \centering
        \includegraphics[width=\textwidth]{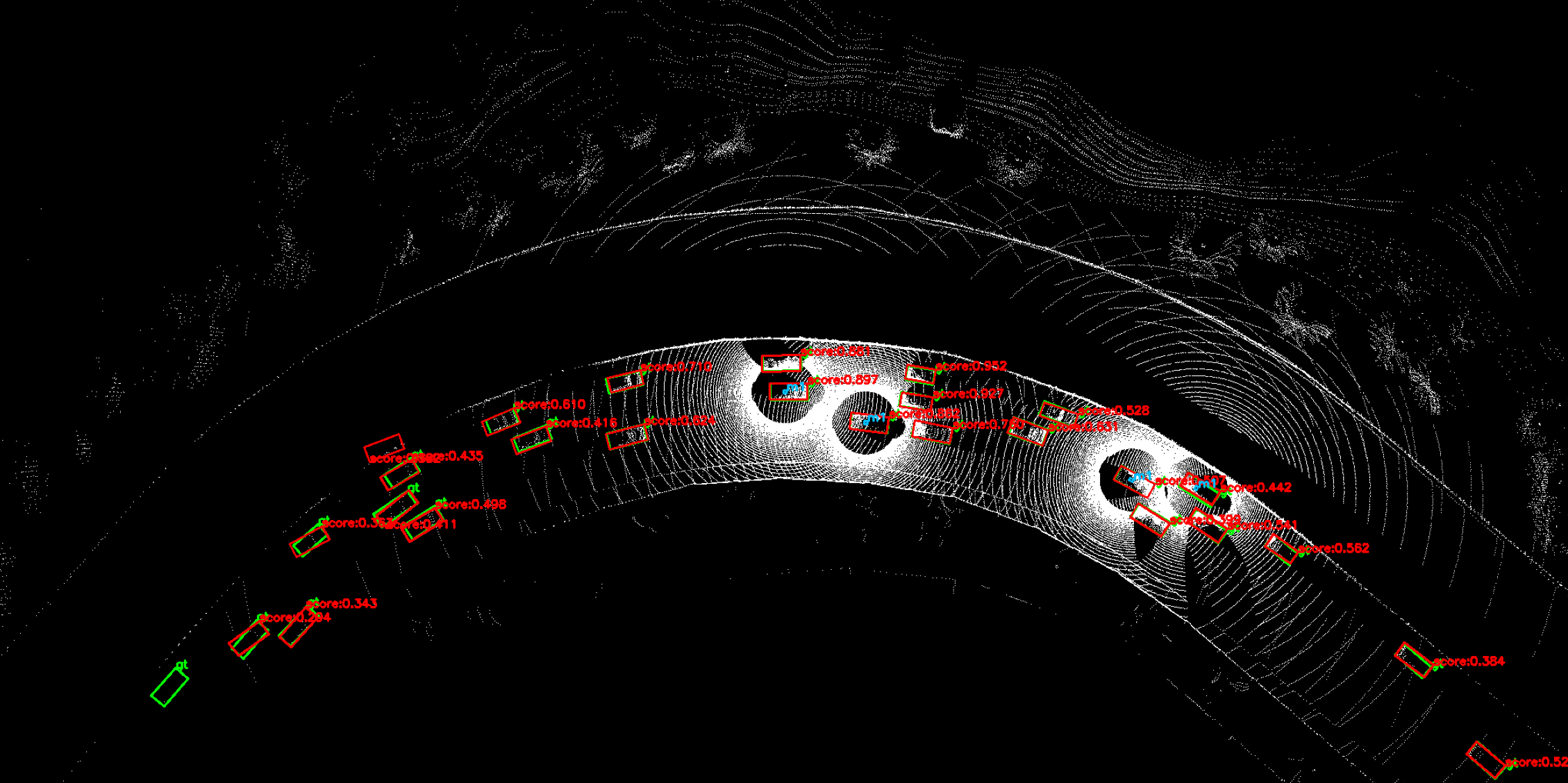}
        \smallskip
        \small OPV2V-LiDAR
    \end{minipage}%
    \hspace{2mm}%
    \begin{minipage}[b]{0.24\textwidth}
        \centering
        \includegraphics[width=\textwidth]{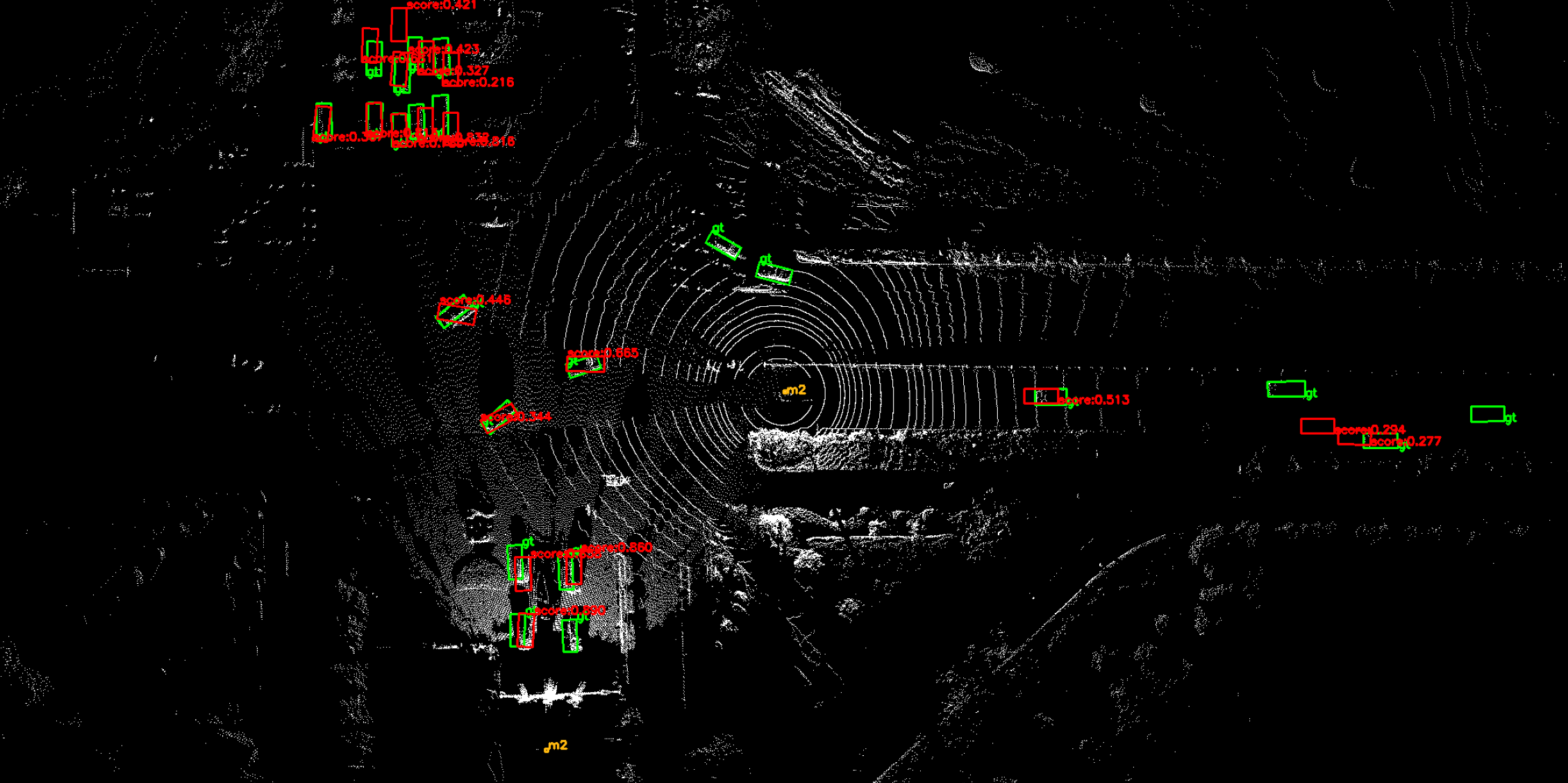}
        \smallskip
        \small DAIR-V2X-Camera
    \end{minipage}%
    \hspace{2mm}%
    \begin{minipage}[b]{0.24\textwidth}
        \centering
        \includegraphics[width=\textwidth]{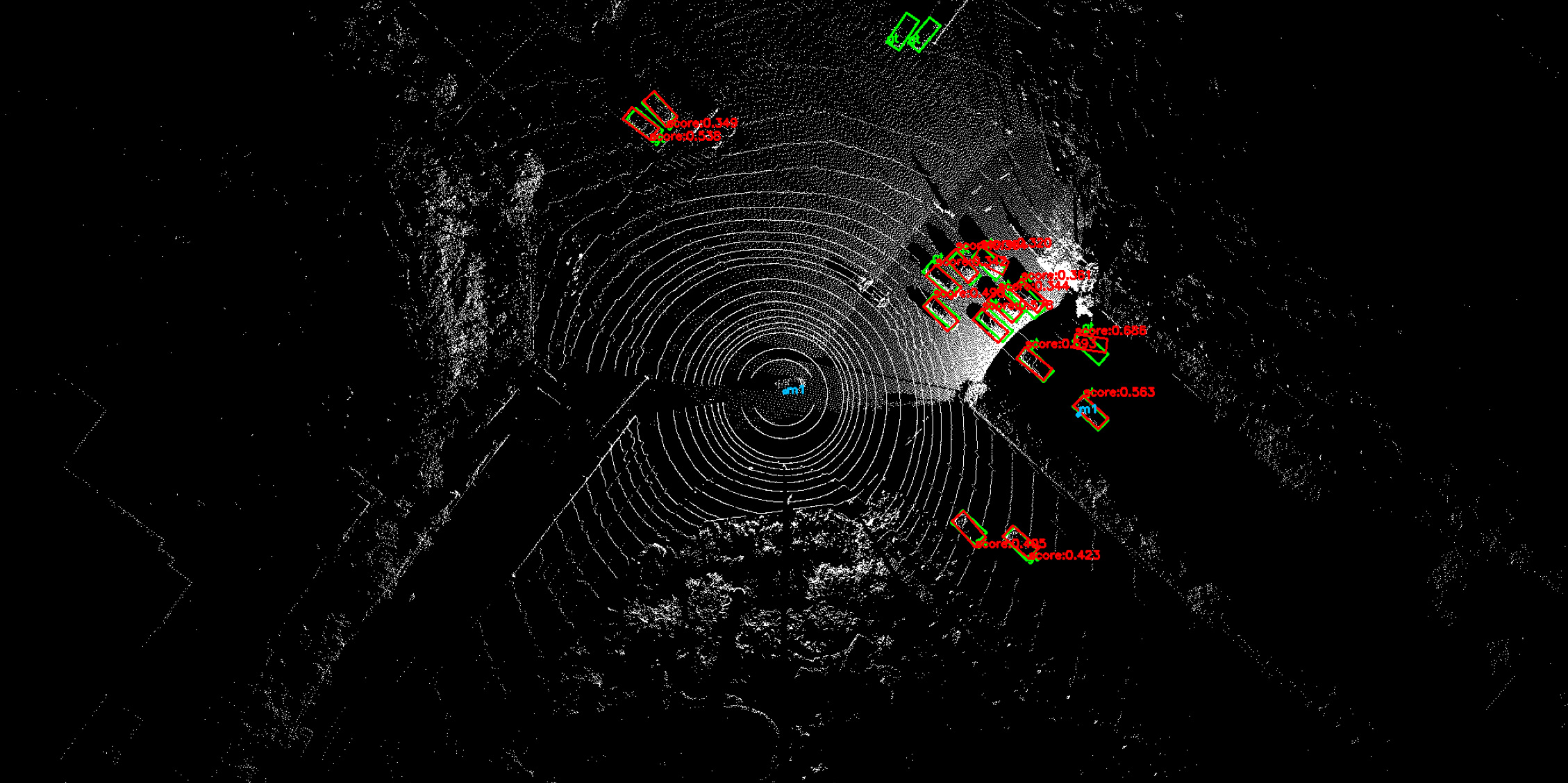}
        \smallskip
        \small DAIR-V2X-LiDAR
    \end{minipage}%

    \caption{Visualization of detection results on the OPV2V and DAIR-V2X datasets under both LiDAR-based and camera-based configurations. Green represents ground truth box, and Red represents predicted box.}
    \label{fig:result_viz}
\end{figure*}

We evaluate the performance of various collaborative 3D object detection models in homogeneous settings using the OPV2V and DAIR-V2X datasets. Figure~\ref{fig:result_viz} presents visualization results for both datasets under camera-based and LiDAR-based conditions, highlighting WaveComm's ability to maintain high recall rates with minimal bandwidth usage by transmitting only the low-frequency component. Additionally, WaveComm exhibits superior accuracy, particularly in LiDAR experiments, where it achieves more precise object detections compared to baselines.

Table~\ref{tab:performance} provides a comprehensive quantitative comparison, demonstrating that our proposed \textbf{\textit{WaveComm}} outperforms state-of-the-art methods, including F-Cooper~\cite{chen2019f}, DiscoNet~\cite{li2021learning}, AttFusion~\cite{xu2022opv2v}, V2X-ViT~\cite{xu2022v2x}, CoBEVT~\cite{xu2022cobevt}, and HM-ViT~\cite{xiang2023hm}, across both LiDAR-based and camera-based homogeneous collaboration tasks. Specifically, on the OPV2V dataset for camera-based perception, WaveComm achieves an AP50 of 0.681 and AP70 of 0.451, surpassing HM-ViT's 0.643 and 0.370 by approximately 5.9\% and 21.9\%, respectively, while reducing communication volume (Comm) from 22.0 to 19.0. In LiDAR-based OPV2V, WaveComm reaches an AP50 of 0.965 and AP70 of 0.926, improving over HM-ViT's 0.950 and 0.873 by 1.6\% and 6.1\%, with Comm lowered from 24.0 to 21.0.

On the DAIR-V2X dataset, for camera-based tasks, WaveComm attains AP30 of 0.274 and AP50 of 0.123, outperforming V2X-ViT's 0.198 and 0.057 by 38.4\% and 115.8\%, respectively, alongside a Comm reduction from 23.0 to 20.0. For LiDAR-based DAIR-V2X, it achieves AP30 of 0.831 and AP50 of 0.790, exceeding HM-ViT's 0.818 and 0.761 by 1.6\% and 3.8\%. These results underscore WaveComm's effectiveness in enhancing detection accuracy while significantly minimizing communication overhead, making it particularly suitable for resource-constrained collaborative perception scenarios.

\subsection{Ablation Studies}

To evaluate the effectiveness and synergy of the proposed method, we conducted three ablation experiments using the baseline model on the DAIR-V2X dataset in the LiDAR-based setting.

\textbf{IDWT and Generator.}
From the results of the comparative experiments in Table~\ref{tab:Ablation1}, it can be seen that the performance of using the wavelet generator to reconstruct is better than that of using only the IDWT module.

\begin{table}[h]
	\caption{Ablation study results of the IDWT and Generator}
        \label{tab:Ablation1}
	\centering
    \resizebox{0.8\linewidth}{!}{
	\begin{tabular}{c|c|c|c}
		\toprule[1pt]
		IDWT & Generator & AP30 $\uparrow$ & AP50 $\uparrow$ \\
		\midrule
		\checkmark &  & 0.826 & 0.783 \\
            \checkmark & \checkmark & 0.831 & 0.790\\
		\bottomrule[1pt]
	\end{tabular}
    }
\end{table}

The generator enhances feature learning by mapping from a latent space and optimizing a loss function to produce richer, more expressive feature representations tailored to the target task, rather than merely restoring input features like the IDWT. While the IDWT's restored features serve as a supervisory signal, the generator does not fully rely on it; instead, by optimizing its loss function alongside this signal, it effectively recovers high-frequency information lost in compression, resulting in more accurate and robust feature representations.

\textbf{Different Component.}
Ablation studies investigate the impact of combined transmission of different frequency components on the performance of collaborative 3D detection. 
In the Base setup, only the low-frequency component is input to the decoder for reconstruction. 
In the Add-Fuse setup, high-frequency components, processed through convolutional and linear layers, are added to the low-frequency component, and the fused result is then input to the decoder. 
In the Concat-Fuse setup, the low-frequency component is concatenated with the processed high-frequency components in the channel dimension, followed by fusion through an additional convolutional layer before being input to the decoder.

\begin{table}[h]
	\caption{Ablation study results of the Different Component}
        \label{tab:Ablation2}
	\centering
    \resizebox{0.7\linewidth}{!}{
	\begin{tabular}{c|c|c}
		\toprule[1pt]
		Component & AP30 $\uparrow$ & AP50 $\uparrow$\\
		\midrule
		Base & 0.826 & 0.783\\
		Add-Fuse & 0.831 & 0.788\\
            Concat-Fuse & 0.829 & 0.783\\
		\bottomrule[1pt]
	\end{tabular}
    }
\end{table}

Table~\ref{tab:Ablation2} shows that both the Add-Fuse and Concat-Fuse methods achieve slight performance improvements over the Base method. However, these two methods introduce higher network complexity and memory usage. To strike a balance between performance improvement and resource efficiency, this paper chooses to adopt the Base method to avoid additional resource overhead while maintaining reasonable performance.

\textbf{Multilevel Wavelet Transform.}
This ablation study evaluates the impact of multi-level wavelet transform on the performance of collaborative 3D detection. As shown in Table~\ref{tab:Ablation3}, varying the number of wavelet transform levels significantly affects detection accuracy. The 1-level transform achieves the highest performance, whereas performance progressively declines with higher levels. This is likely due to increased feature loss caused by excessive decomposition.

\begin{table}[h]
	\caption{Ablation study results of the Multilevel Wavelet Transform}
        \label{tab:Ablation3}
	\centering
    \resizebox{0.8\linewidth}{!}{
	\begin{tabular}{c|c|c|c}
		\toprule[1pt]
		 Multilevel  & AP30 $\uparrow$ & AP50 $\uparrow$ & Comm\\
		\midrule
		1-level & 0.831 & 0.790 & 20.0\\
		2-level & 0.825 & 0.768 & 18.0\\
            3-level & 0.801 & 0.721 & 18.0\\
		\bottomrule[1pt]
	\end{tabular}
    }
\end{table}

\section{CONCLUSIONS}

In this paper, we propose WaveComm, a lightweight collaborative perception framework leveraging wavelet feature distillation to address excessive communication overhead. 
We design a wavelet-based compression–reconstruction mechanism: feature maps are decomposed into low- and high-frequency components, only compact low-frequency parts are transmitted, and a lightweight generator reconstructs missing details at the receiver. 
A multi-loss optimization further improves reconstruction fidelity under low communication cost. 
Experiments on multiple datasets show that WaveComm achieves strong detection accuracy with significantly reduced bandwidth. 
Future work will study WaveComm's robustness under more realistic V2X conditions, including heterogeneous agents and unreliable communication. 

\addtolength{\textheight}{0cm}   








\bibliographystyle{IEEEtran}
\balance
\bibliography{ref}

@article{arnold2020cooperative,
  title={Cooperative perception for 3D object detection in driving scenarios using infrastructure sensors},
  author={Arnold, Eduardo and Dianati, Mehrdad and De Temple, Robert and Fallah, Saber},
  journal={IEEE Transactions on Intelligent Transportation Systems},
  volume={23},
  number={3},
  pages={1852--1864},
  year={2020},
  publisher={IEEE}
}

@inproceedings{wang2020v2vnet,
  title={V2vnet: Vehicle-to-vehicle communication for joint perception and prediction},
  author={Wang, Tsun-Hsuan and Manivasagam, Sivabalan and Liang, Ming and Yang, Bin and Zeng, Wenyuan and Urtasun, Raquel},
  booktitle={European conference on computer vision},
  pages={605--621},
  year={2020},
  organization={Springer}
}

@article{liu2023towards,
  title={Towards vehicle-to-everything autonomous driving: A survey on collaborative perception},
  author={Liu, Si and Gao, Chen and Chen, Yuan and Peng, Xingyu and Kong, Xianghao and Wang, Kun and Xu, Runsheng and Jiang, Wentao and Xiang, Hao and Ma, Jiaqi and others},
  journal={arXiv preprint arXiv:2308.16714},
  year={2023}
}

@article{lu2024extensible,
  title={An extensible framework for open heterogeneous collaborative perception},
  author={Lu, Yifan and Hu, Yue and Zhong, Yiqi and Wang, Dequan and Wang, Yanfeng and Chen, Siheng},
  journal={arXiv preprint arXiv:2401.13964},
  year={2024}
}

@article{hu2022where2comm,
  title={Where2comm: Communication-efficient collaborative perception via spatial confidence maps},
  author={Hu, Yue and Fang, Shaoheng and Lei, Zixing and Zhong, Yiqi and Chen, Siheng},
  journal={Advances in neural information processing systems},
  volume={35},
  pages={4874--4886},
  year={2022}
}

@inproceedings{liu2020when2com,
  title={When2com: Multi-agent perception via communication graph grouping},
  author={Liu, Yen-Cheng and Tian, Junjiao and Glaser, Nathaniel and Kira, Zsolt},
  booktitle={Proceedings of the IEEE/CVF Conference on computer vision and pattern recognition},
  pages={4106--4115},
  year={2020}
}

@article{yu2025which2comm,
  title={Which2comm: An efficient collaborative perception framework for 3d object detection},
  author={Yu, Duanrui and You, Jing and Pei, Xin and Qu, Anqi and Wang, Dingyu and Jia, Shaocheng},
  journal={arXiv preprint arXiv:2503.17175},
  year={2025}
}

@inproceedings{xu2022opv2v,
  title={Opv2v: An open benchmark dataset and fusion pipeline for perception with vehicle-to-vehicle communication},
  author={Xu, Runsheng and Xiang, Hao and Xia, Xin and Han, Xu and Li, Jinlong and Ma, Jiaqi},
  booktitle={2022 International Conference on Robotics and Automation (ICRA)},
  pages={2583--2589},
  year={2022},
  organization={IEEE}
}

@inproceedings{yu2022dair,
  title={Dair-v2x: A large-scale dataset for vehicle-infrastructure cooperative 3d object detection},
  author={Yu, Haibao and Luo, Yizhen and Shu, Mao and Huo, Yiyi and Yang, Zebang and Shi, Yifeng and Guo, Zhenglong and Li, Hanyu and Hu, Xing and Yuan, Jirui and others},
  booktitle={Proceedings of the IEEE/CVF Conference on Computer Vision and Pattern Recognition},
  pages={21361--21370},
  year={2022}
}

@inproceedings{xu2023v2v4real,
  title={V2v4real: A real-world large-scale dataset for vehicle-to-vehicle cooperative perception},
  author={Xu, Runsheng and Xia, Xin and Li, Jinlong and Li, Hanzhao and Zhang, Shuo and Tu, Zhengzhong and Meng, Zonglin and Xiang, Hao and Dong, Xiaoyu and Song, Rui and others},
  booktitle={Proceedings of the IEEE/CVF conference on computer vision and pattern recognition},
  pages={13712--13722},
  year={2023}
}

@inproceedings{xu2021opencda,
  title={Opencda: an open cooperative driving automation framework integrated with co-simulation},
  author={Xu, Runsheng and Guo, Yi and Han, Xu and Xia, Xin and Xiang, Hao and Ma, Jiaqi},
  booktitle={2021 IEEE International Intelligent Transportation Systems Conference (ITSC)},
  pages={1155--1162},
  year={2021},
  organization={IEEE}
}

@inproceedings{dosovitskiy2017carla,
  title={CARLA: An open urban driving simulator},
  author={Dosovitskiy, Alexey and Ros, German and Codevilla, Felipe and Lopez, Antonio and Koltun, Vladlen},
  booktitle={Conference on robot learning},
  pages={1--16},
  year={2017},
  organization={PMLR}
}

@article{wang2025cocmt,
  title={Cocmt: Communication-efficient cross-modal transformer for collaborative perception},
  author={Wang, Rujia and Gao, Xiangbo and Xiang, Hao and Xu, Runsheng and Tu, Zhengzhong},
  journal={arXiv preprint arXiv:2503.13504},
  year={2025}
}

@inproceedings{hu2024communication,
  title={Communication-efficient collaborative perception via information filling with codebook},
  author={Hu, Yue and Peng, Juntong and Liu, Sifei and Ge, Junhao and Liu, Si and Chen, Siheng},
  booktitle={Proceedings of the IEEE/CVF Conference on Computer Vision and Pattern Recognition},
  pages={15481--15490},
  year={2024}
}

@inproceedings{zhang2024ermvp,
  title={Ermvp: Communication-efficient and collaboration-robust multi-vehicle perception in challenging environments},
  author={Zhang, Jingyu and Yang, Kun and Wang, Yilei and Wang, Hanqi and Sun, Peng and Song, Liang},
  booktitle={Proceedings of the IEEE/CVF Conference on Computer Vision and Pattern Recognition},
  pages={12575--12584},
  year={2024}
}

@inproceedings{jin2025bandwidth,
  title={Bandwidth-Efficient Communication Modelling for Autonomous Vehicle Collaborative Perception},
  author={Jin, Dinghao and Zeng, Yuan and Gong, Yi},
  booktitle={2025 IEEE/CVF Winter Conference on Applications of Computer Vision (WACV)},
  pages={6146--6155},
  year={2025},
  organization={IEEE}
}

@article{zhang2025fast2comm,
  title={Fast2comm: Collaborative perception combined with prior knowledge},
  author={Zhang, Zhengbin and Wu, Yan and Zhang, Hongkun},
  journal={arXiv preprint arXiv:2505.00740},
  year={2025}
}

@article{wang2025coopdetr,
  title={CoopDETR: A unified cooperative perception framework for 3D detection via object query},
  author={Wang, Zhe and Xu, Shaocong and Zhuang, Xucai and Xu, Tongda and Wang, Yan and Liu, Jingjing and Chen, Yilun and Zhang, Ya-Qin},
  journal={arXiv preprint arXiv:2502.19313},
  year={2025}
}

@inproceedings{chen2019f,
  title={F-cooper: Feature based cooperative perception for autonomous vehicle edge computing system using 3D point clouds},
  author={Chen, Qi and Ma, Xu and Tang, Sihai and Guo, Jingda and Yang, Qing and Fu, Song},
  booktitle={Proceedings of the 4th ACM/IEEE Symposium on Edge Computing},
  pages={88--100},
  year={2019}
}

@article{li2021learning,
  title={Learning distilled collaboration graph for multi-agent perception},
  author={Li, Yiming and Ren, Shunli and Wu, Pengxiang and Chen, Siheng and Feng, Chen and Zhang, Wenjun},
  journal={Advances in Neural Information Processing Systems},
  volume={34},
  pages={29541--29552},
  year={2021}
}

@inproceedings{xu2022v2x,
  title={V2x-vit: Vehicle-to-everything cooperative perception with vision transformer},
  author={Xu, Runsheng and Xiang, Hao and Tu, Zhengzhong and Xia, Xin and Yang, Ming-Hsuan and Ma, Jiaqi},
  booktitle={European conference on computer vision},
  pages={107--124},
  year={2022},
  organization={Springer}
}

@article{xu2022cobevt,
  title={CoBEVT: Cooperative bird's eye view semantic segmentation with sparse transformers},
  author={Xu, Runsheng and Tu, Zhengzhong and Xiang, Hao and Shao, Wei and Zhou, Bolei and Ma, Jiaqi},
  journal={arXiv preprint arXiv:2207.02202},
  year={2022}
}

@inproceedings{xiang2023hm,
  title={HM-ViT: Hetero-modal vehicle-to-vehicle cooperative perception with vision transformer},
  author={Xiang, Hao and Xu, Runsheng and Ma, Jiaqi},
  booktitle={Proceedings of the IEEE/CVF international conference on computer vision},
  pages={284--295},
  year={2023}
}

@inproceedings{yang2023how2comm,
 title={How2comm: Communication-efficient and collaboration-pragmatic multi-agent perception},
 author={Yang, Dingkang and Yang, Kun and Wang, Yuzheng and Liu, Jing and Xu, Zhi and Yin, Rongbin and Zhai, Peng and Zhang, Lihua},
 booktitle={Thirty-seventh Conference on Neural Information Processing Systems (NeurIPS)},
 year={2023}
}

@article{finder2022wavelet,
  title={Wavelet feature maps compression for image-to-image CNNs},
  author={Finder, Shahaf E and Zohav, Yair and Ashkenazi, Maor and Treister, Eran},
  journal={Advances in Neural Information Processing Systems},
  volume={35},
  pages={20592--20606},
  year={2022}
}

@article{xiang2024remote,
  title={Remote sensing image compression based on high-frequency and low-frequency components},
  author={Xiang, Shao and Liang, Qiaokang},
  journal={IEEE Transactions on Geoscience and Remote Sensing},
  volume={62},
  pages={1--15},
  year={2024},
  publisher={IEEE}
}

@article{song2024high,
  title={High frequency matters: Uncertainty guided image compression with wavelet diffusion},
  author={Song, Juan and He, Jiaxiang and Yang, Lijie and Feng, Mingtao and Wang, Keyan},
  journal={arXiv preprint arXiv:2407.12538},
  year={2024}
}

@inproceedings{su2024makes,
  title={What makes good collaborative views? contrastive mutual information maximization for multi-agent perception},
  author={Su, Wanfang and Chen, Lixing and Bai, Yang and Lin, Xi and Li, Gaolei and Qu, Zhe and Zhou, Pan},
  booktitle={Proceedings of the AAAI conference on artificial intelligence},
  volume={38},
  number={16},
  pages={17550--17558},
  year={2024}
}

@article{yu2023flow,
  title={Flow-based feature fusion for vehicle-infrastructure cooperative 3d object detection},
  author={Yu, Haibao and Tang, Yingjuan and Xie, Enze and Mao, Jilei and Luo, Ping and Nie, Zaiqing},
  journal={Advances in Neural Information Processing Systems},
  volume={36},
  pages={34493--34503},
  year={2023}
}

@article{gal2021swagan,
  title={Swagan: A style-based wavelet-driven generative model},
  author={Gal, Rinon and Hochberg, Dana Cohen and Bermano, Amit and Cohen-Or, Daniel},
  journal={ACM Transactions on Graphics (TOG)},
  volume={40},
  number={4},
  pages={1--11},
  year={2021},
  publisher={ACM New York, NY, USA}
}

@inproceedings{huang2017wavelet,
  title={Wavelet-srnet: A wavelet-based cnn for multi-scale face super resolution},
  author={Huang, Huaibo and He, Ran and Sun, Zhenan and Tan, Tieniu},
  booktitle={Proceedings of the IEEE international conference on computer vision},
  pages={1689--1697},
  year={2017}
}

@article{duan2017sar,
  title={SAR image segmentation based on convolutional-wavelet neural network and Markov random field},
  author={Duan, Yiping and Liu, Fang and Jiao, Licheng and Zhao, Peng and Zhang, Lu},
  journal={Pattern Recognition},
  volume={64},
  pages={255--267},
  year={2017},
  publisher={Elsevier}
}

@inproceedings{williams2018wavelet,
  title={Wavelet pooling for convolutional neural networks},
  author={Williams, Travis and Li, Robert},
  booktitle={International conference on learning representations},
  year={2018}
}

@inproceedings{wolter2020neural,
  title={Neural network compression via learnable wavelet transforms},
  author={Wolter, Moritz and Lin, Shaohui and Yao, Angela},
  booktitle={International Conference on Artificial Neural Networks},
  pages={39--51},
  year={2020},
  organization={Springer}
}

@inproceedings{rho2023masked,
  title={Masked wavelet representation for compact neural radiance fields},
  author={Rho, Daniel and Lee, Byeonghyeon and Nam, Seungtae and Lee, Joo Chan and Ko, Jong Hwan and Park, Eunbyung},
  booktitle={Proceedings of the IEEE/CVF Conference on Computer Vision and Pattern Recognition},
  pages={20680--20690},
  year={2023}
}

@inproceedings{isola2017image,
  title={Image-to-image translation with conditional adversarial networks},
  author={Isola, Phillip and Zhu, Jun-Yan and Zhou, Tinghui and Efros, Alexei A},
  booktitle={Proceedings of the IEEE conference on computer vision and pattern recognition},
  pages={1125--1134},
  year={2017}
}

@article{mao2024diffcp,
  title={DiffCP: Ultra-Low Bit Collaborative Perception via Diffusion Model},
  author={Mao, Ruiqing and Wu, Haotian and Jia, Yukuan and Nan, Zhaojun and Sun, Yuxuan and Zhou, Sheng and G{\"u}nd{\"u}z, Deniz and Niu, Zhisheng},
  journal={arXiv preprint arXiv:2409.19592},
  year={2024}
}

@inproceedings{lang2019pointpillars,
  title={Pointpillars: Fast encoders for object detection from point clouds},
  author={Lang, Alex H and Vora, Sourabh and Caesar, Holger and Zhou, Lubing and Yang, Jiong and Beijbom, Oscar},
  booktitle={Proceedings of the IEEE/CVF conference on computer vision and pattern recognition},
  pages={12697--12705},
  year={2019}
}

\end{document}